\documentclass[graybox]{svproc} 

\usepackage{type1cm}        
\usepackage{pgfplots}
\usepackage{amsmath}
\usepackage{makeidx}         
\usepackage{multicol}        
\usepackage[bottom]{footmisc}

\usepackage{subcaption}
\usepackage{mathtools}
\DeclarePairedDelimiter{\ceil}{\lceil}{\rceil}
\usepackage{newtxtext}       %
\usepackage[varvw]{newtxmath}       
\usepackage{caption}
\usepackage{graphicx}

\newsavebox{\measurebox}

\makeindex             

\begin{document}

\title{AutoDrone: Shortest Optimized Obstacle-Free Path Planning for Autonomous Drones}
\author{Prithwish Jana\inst{1} \and Debasish Jana\inst{2}}
\institute{Member, IEEE \\ Indian Institute of Technology Kharagpur, India\\
	 \email{pjana@ieee.org}
\and 
Senior Member, IEEE \\ TEOCO Software Pvt. Ltd., Kolkata, India\\
 \email{debasishj@gmail.com}}
%
%

\maketitle

\begin{abstract}
	With technological advancement, drone has emerged as unmanned aerial vehicle that can be controlled by humans to fly or reach a destination. This may be autonomous as well, where the drone itself is intelligent enough to find a shortest obstacle-free path to reach the destination from a designated source. Be it a planned smart city or even a wreckage site affected by natural calamity, we may imagine the buildings, any surface-erected structure or other blockage as obstacles for the drone to fly in a straight line-of-sight path. To address such path-planning of drones, the bird's eye-view of the whole landscape is first transformed to a graph of grid-cells, where some are occupied to indicate the obstacles and some are free to indicate the free path. We propose a method to find out the shortest obstacle-free path in the coordinate system guided by GPS. The autonomous drone (AutoDrone) will thus be able to move from one place to another along the shortest path, without colliding into hindrances, while traveling in a two-dimensional space. Heuristics to extend this to long journeys and 3D space are also elaborated. Our approach can be especially beneficial in rescue operations and fast delivery or pick-up in an energy-efficient way, where our algorithm will help in finding out the shortest path and angle along which it should fly. Experiments are done on different scenarios of map layouts and obstacle positions to understand the shortest feasible route, computed by the autonomous drone.

\end{abstract}

\section{Introduction}
\label{intro}
Drone \cite{cauchard2015drone} is the common name of an aerial vehicle that is unmanned, has no pilot to make them fly, no cabin crew to serve on air and no passenger to fly along. This was originally thought to serve aerospace applications or military use. Now-a-days, the usage has been in personal use as well as for reaching out to places where human may find difficulty to reach. Drone needs to be operated remotely by a human on ground through a screen-enabled remote-control device within a short range. Drones promise to be the mode-of-delivery in India’s future smart city project. They can be used to deliver things to the desired destination. In case of autonomous drone (AutoDrone), an intelligent drone may have to fly without being remotely controlled by a human to follow order. Rather, a drone may be used to fly on their own for traversing a trajectory towards a desired destination. It may rely on artificial intelligence backbone to imitate or act like an efficient human pilot. This may be especially useful in search or relief operations across rough terrains or inaccessible corners of monumental ruins in an area devastated by disasters like earthquake or tsunami. In such places, time is crucial and radio-signals may be impenetrable through impassable frontiers or unassailable mountains -- as such, autonomous systems is need of the hour.

To take control of in-flight situations, the decision engine within a drone may be embedded. This decision engine must be intelligent enough to interpret raw-sensor data meaningfully. With camera-guided real-time vision and interpretation of the path and obstacle, drone can fly to destination using shortest hindrance-free path to travel. Thus, the intelligent engine within the drone should be able to choose a shortest travel-trajectory when there is no preset list of defined paths (that is, drone has no prior information about the terrain it is going to enter), alongside avoiding the obstacles. If we map the whole terrain to a rectangular grid of certain precision, where some grid-cells represent obstacles and the drone should avoid collision with those, then we can reduce the set-up of the problem to a two-dimensional Euclidean planar graph~\cite{bounceur2018boundaries}. Ideally, there could be an unobstructed direct route, connecting source and destination. Using GPS-guided coordinate system, an autonomous intelligent drone can easily find out the shortest path and angle along which it should drive. In reality, there could be obstacles as buildings, monuments, elevated foot-paths, electricity towers, non-flyable areas and on-air hindrances that may creep in dynamically. 

Our work focuses on finding out the shortest path without obstacle from a source to a chosen destination. The implementation testbed for simulation is written in C++~\cite{jana2014c++} on Ubuntu platform. The rest of the paper is organized as follows: Section~\ref{relwork} jots down the related work in this area. Section~\ref{proposed} elaborates the proposed method. This follows simulation experimental results in Section~\ref{results}. Finally we conclude in Section~\ref{conclusion}.

\section{Related Work}
\label{relwork}
From a graph-theoretic \cite{deo2017graph} perspective, a directed or undirected connected graph consist of several nodes (or vertices) and a set of edges that link the nodes. The shortest path from a source node to a target node is the route joining them via edges such that, the number of edges (in unweighted graph) or the sum of the respective weights of the connecting edges (in weighted graph) is minimal. Similarly, a map of connecting roads may be represented as a Manhattan-graph of intersections (vertices) and the road connecting two intersections (edges). Finding the shortest path in such a simplified road map can be essentially done by reducing it to graph-theoretic shortest path probing. As such, each edge is weighed in accordance with the length of the road segment that it represents.

Gallo et al.~\cite{gallo1988shortest} have listed eight popular algorithms that show the way to solve the shortest path problem on a tree or a directed graph. Dijkstra's shortest path algorithm has been explained by Johnson~\cite{johnson1973note}. There are many algorithms for finding the shortest path in a graph, e.g. the popular ones include Bellman-Ford's algorithm~\cite{sulaiman2018bellman}, Floyd-Warshall algorithm~\cite{burfield2013floyd} and Dijkstra's algorithm~\cite{chen2003dijkstra}. Bellman-Ford algorithm is used to find shortest paths in a graph, from a particular source vertex to all other vertices. Floyd-Washall algorithm finds the shortest paths between all the existing pairs of vertices in a graph, where each connecting edge present in the graph has some non-zero weight, positive or negative. Dijkstra's algorithm deals with single-source shortest path, and as such, is more appropriate in our problem domain. But for our problem set-up, this cannot be applied directly as-is in the original form directly because, paths are not well-defined in an Euclidean plane and there may be obstacles blocking the line-of-sight paths. 

Chai et al.~\cite{chai2020aerial} predict spatial context through computation of distance map between a designated pixel to its spatially nearest boundary of an object. We may envisage a road-map of connecting aerial routes and erected buildings as a graph of vertices (identified objects) and edges (connecting aerial connections) to signify \textit{connectivity}. Thus, presence of an edge between two nodes means a drone can fly among these two nodes aerially in Euclidean space. Mutherjee at el.~\cite{mukherjee2020two} worked on semantic segmentation on images to do an effective annotation at pixel-level. Thus a roadmap with erected buildings and monuments of a city-scape or village-scape can be semantically segmented to a map, and eventually to a graph in Euclidean space. And the task for the AutoDrone boils down to the task to find the single-source shortest path in a directed graph in Euclidean space. We may assume that in an Euclidean graph, the estimated weight of each connecting edge can be provided by the Euclidean distance between its two points.

Chakrabarty et al.~\cite{chakrabarty2019real} demonstrated the interpretation of autonomous traversing of a drone from a designated source $A$ to a designated target point $B$, when there is no obstacle in between. As such in reality, it may not be completely aerially clear between two points but, it may be among few other intermediate nodes as multiple hops where each hop is going through some aerially clear edge. Within the Unmanned Aircraft System Traffic Management (UTM) ecosystem, a drone is able to fly across an aerially clear path through multiple hops. Jana~\cite{pj2020ysc} has argued that shortest path traversal needs to be through `visually clear' path without any obstacles, be it for autonomous driver-less car or pilot-less aircraft like drone. According to Koch et al.~\cite{koch2019automatic}, while mapping mostly a landscape scene, be it spacious or flat scenes, flight planning for drone can be thought of simple grid-like structure where drones can follow two-dimensional patterns of nodes and edges within 2D space. Sakurai et al.~\cite{sakurai2021path} proposed a spiking neural network (SNN) algorithm to compute plan for appropriate path having several moving obstacles around target agents that itself is attempting to minimize its own traversal path to the destination. In practical scenarios, there could be many other moving obstacles. As for example, this may include other cars for a driver-less autonomous car and human or building obstacles for a moving AutoDrone or robot. In such situations, getting an effective shortest path is not a trivial task, because the dynamic behavior of the obstacles may not be known or there could be a large number of other candidate paths to choose from. But from a drone's perspective that flies at a certain height, obstacles are more or less static except for birds and other drones. So, in this work we tackle such obstacles and propose a path-planning algorithm in 3D space that can maneuver the drone from its current position to an intended destination. 

\section{Proposed Methodology}
\label{proposed}

In this section, we first list out perception techniques to map surrounding neighborhood to a 2D layout. We then provide the problem description and describe our proposed method of path-planning. Finally, we transform this to 3D Euclidean space to accomodate source-destination pairs at different altitudes.

\subsection{Perception and Real-time Formation of Maps}
\label{perception}
At the starting location, the drone can first move vertically upwards to a high altitude from where almost all obstacles would be at a lower altitude. From that position, the drone's camera can capture high-resolution RGB images that represents a bird's eye-view of the extended region surrounding the drone. Fast semantic segmentation on RGB images through fully-convolutional network~\cite{wu2018automatic} or superpixel-based classification~\cite{mukherjee2020two} can label each pixel to their corresponding class as building, tree, road, tower, etc. Further, panoptic segmentation~\cite{kirillov2019panoptic} separates different instances of the same object (e.g. two different buildings). In instances of fog, smog or any other technical glitch, when captured RGB image is blurry with lot of noises,  binarization~\cite{jana2017fuzzy}\cite{jana2017handwritten} techniques can be applied on fuzzy image segmentation results to get a cleaner per-pixel binary classification of obstacle/not-obstacle. All these, individually or in combination, give an accurate 2D aerial map delineating the positions of obstacles. In addition, airborne 3D LiDAR~\cite{rajagopal2020deeposm} fitted on the drone can proffer depth readings that can inversely signify the height of the various obstacles. Fusing such multi-sensor information, by methods like bimodal learning~\cite{parajuli2018fusion} from RGB-D information or unified co-attention networks~\cite{zhou2020rgb} can augment the spatial maps with depth information. Global Positioning System (GPS) helps in identifying the intended destination coordinate in the captured image. As such, the drones can determine at which height it should move depending on at which height it would encounter more percentage of open-area without hindrances.

\subsection{Problem Design}
\label{probDesign}
\begin{figure}[!h]
	\centering
	\fbox{
	\includegraphics[width=0.7\textwidth]{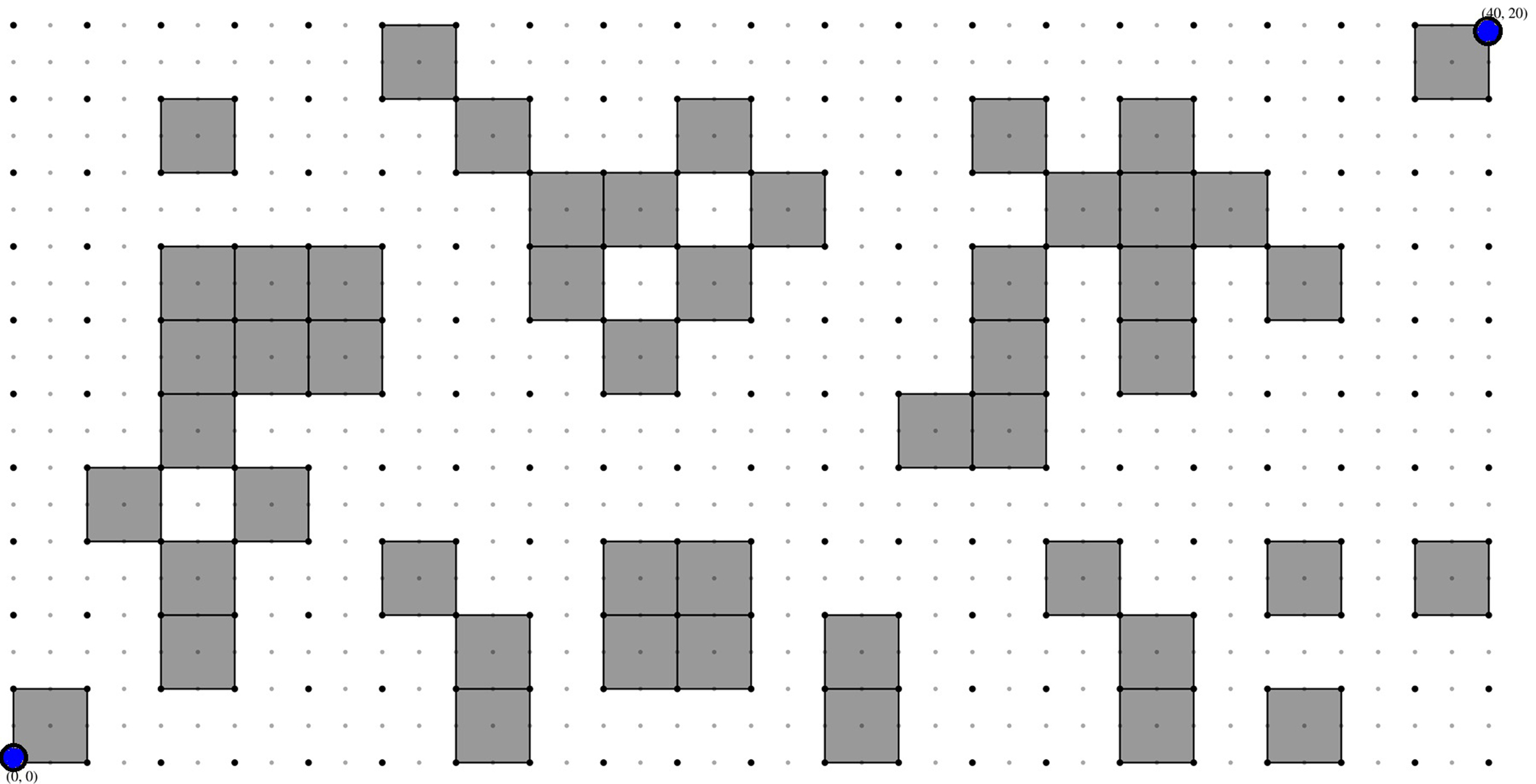}}
	\caption{Example of 2D grid formed from a 10m $\times$ 20m area. Each cell (denoted within black dot corners) is 1m $\times$ 1m. Shaded cells denote obstacles. As such, the straight line joining source at the bottom-left to destination at the top-right (shown as blue circles) is not obstacle-free and drone has to find a path that avoids such obstacles}
	\label{fig:gridExample}
\end{figure}

We consider that the drone locomotes across an area defined by a two-dimensional region of size $R$ meters $\times$ $C$ meters, as exemplified in Figure~\ref{fig:gridExample}. There are a set of destinations $D=\{d_1, d_2, \cdots\}$ where the drone needs to deliver items. Additionally, there are obstacles spread throughout the area that hinders the straight line-of-sight path from the drone base-station to each of the destinations. First, with regard to the precision ($p$ metres) of motion desired, we discretize the 2D space into a $\ceil[\big]{\frac{R}{p}}\times \ceil[\big]{\frac{C}{p}}$ grid of ($p$ metres $\times$ $p$ metres) unit squares. Identifying the lower-leftmost point of the grid as the origin $(0,0)$, individual destination locations are defined by 2D Cartesian coordinates $d_i=(x_i,y_i)$. Further a drone's obstacles can be buildings or trees that occupy a polygonal area (convex or concave), when seen through bird's eye view. At the height a drone is set to fly, each obstacle may look arbitrary-shaped. So, the drone should avoid the region defined by the convex hull of that obstacle, to avoid collisions. Thus, if some grid-cell overlaps with the convex hull to such polygons it is considered as an \textit{obstacle-grid-cell}. In our design, there are no partially-obstacled grid-cells i.e., each cell is either wholly occupied by an obstacle or it is not. Neighboring obstacle-grid-cells together can make up a larger obstacle of arbitrary rectilinear-polygonal shape.

\subsection{Destination to Base-Station Shortest Path}
\label{destBaseShortest}
We intend to find the shortest source-destination path from the drone's base-station to a destination $d_i$. Now, when there are no obstacles blocking the line-of-sight between source and the destination, the shortest Euclidean path will be a straight line joining them. But, a drone may have to deviate from this line-of-sight path when obstacles adorn the drone's flying area. Instead of going straight, the drone would have to change its direction at specific coordinates to avoid collision and reach safe. In this paper, we provide a fast and efficient approach to compute an optimized source-destination shortest path that a drone would take, in order to avoid collision with obstacles. To solve this efficiently, we take a computational geometry based approach, as described in the following paragraphs. 

\paragraph{\textbf{Obstacle Graph Formation.}} As we said before, the drone has to shift the direction at specific coordinates (\textit{points of deflection}). And for the drone to follow the shortest path, it should change directions not just anywhere but only at obstacle corners. This is because, the shortest source-destination path can be imagined as a tight string between source and destination that when in its fully contracted position, will only touch some of the obstacle vertices that does not allow it to contract further. It may be noted here that the shortest path may not always be convex in nature. So, we firstly form a graph known as the \textit{obstacle graph} $G_O$ whose vertices are the set of all corner-points corresponding to obstacle-grid-cells. If the shortest source-destination path contains any position from which the drone has to deflect its direction, it will definitely be one among the vertices in $G_O$. Further, any vertex in $G_O$ may be shared among one, two, three or four obstacle-grid-cells. The vertices that are shared by four obstacle-grid-cells are completely interior to a bigger obstacle formed by the neighboring obstacle-grid-cells. And therefore, they cannot be points of deflection. We mark these special vertices and may safely exclude them in the visibility graph computation (explained in the subsequent paragraph) because these vertices cannot be `seen' from any other vertex. This way, we reduce the time to probe whether line-of-sight between two points is obstructed or not. In $G_O$, we add an edge between obstacle vertices $u,v\in G_O$ if edge $(u,v)$ is an obstacle edge i.e. a bounding edge of an obstacle-grid-cell.

\paragraph{\textbf{Visibility Graph Formation.}} Subsequently, a \textit{visibility graph} is formed from here. It is an Euclidean graph where vertices connected by edges are mutually visible from one another i.e. their straight-line path is not blocked by any obstacle. In this case, the drone can move unobstructed amongst these two points and the path will also be shortest Euclidean path. On the contrary, the drone may need to change directions in-between when there does not exist an edge between two vertices. As we mentioned before, a vertex that lies on the shortest path from source $S$ to destination $D$ includes $S$, $D$ and one or more unmarked obstacle-vertices $v \in G_O$, and nothing else. As such, we improve upon the overall computation time by not probing `visibility' status among all ${\ceil[\big]{\frac{R}{p}}\times \ceil[\big]{\frac{C}{p}} \choose 2}$ pairs, but only for each pair of vertices in $G_O$, which in general is much smaller than the former. 

\paragraph{\textbf{Checking Visibility between Vertex-Pairs.}} Now that we fixed the vertices for our visibility graph, we need to add undirected edges between the pairs that can mutually `see' each other. Such edges will be \textit{intermediate subpaths} of the overall S-D shortest path. Taking each vertex on the visibility graph as the intermediate source ($S_{in}$), we systematically check if the other vertices placed to its right (intermediate destination, $D_{in}$) are visible from the former or not. So for each $S_{in}$, we consider a shifted coordinate system with $S_{in}$ at origin $(0,0)$ and check only those $D_{in}$ that lie in the first and fourth quadrant. This will complete visibility computation for potential intermediate subpaths having slope between $-90^{\circ}$ to $90^{\circ}$. $D_{in}$ in the second and third quadrant i.e. potential intermediate subpaths having slope between $90^{\circ}$ to $270^{\circ}$ will be automatically completed by symmetricity of undirected edges. In particular, four distinct mutually exclusive cases can arise when checking visibility between $S_{in}$ and $D_{in}$ (all direction specifications are with respect to a 2D plane):

\begin{enumerate}
    \item \textit{$D_{in}$ is vertically above or below $S_{in}$}. A drone can fly along edges of an obstacle except those that are contained completely within an obstacle. These latter edges can be computed in linear time from the obstacle-graph when we find those edges that are shared by two obstacles. We call these \textit{blocking-edge}s because when it is present between two points, these edges block their mutual line-of-sight and there is no visibility of one from another. So, if some vertical blocking-edge overlaps (fully or partially) the vertical-line joining $S_{in}$ and $D_{in}$, then they are not mutually intervisible. If no such blocking-edge is there, $S_{in}$ and $D_{in}$ are intervisible.
    \item \textit{$D_{in}$ is horizontally rightwards to $S_{in}$}. Similar to the previous case, we check from $G_O$ whether there exists some horizontal blocking-edge that overlaps (fully or partially) the horizontal-line joining $S_{in}$ and $D_{in}$. If not, they are are mutually intervisible.
    \item \textit{$D_{in}$ is diagonally rightwards to $S_{in}$ i.e. makes a $45^{\circ}$ or $-45^{\circ}$ angle}. From obstacle graph $G_O$, we check if the line joining $S_{in}$ and $D_{in}$ contains any obstacle vertices which is also the left-bottom or left-top corner of an obstacle-grid-cell. If not, they are are mutually intervisible.
    \item \textit{$D_{in}$ is in a generic position w.r.t $S_{in}$ other than the previous cases}. To check whether two vertices $S_{in}$ and $D_{in}$ are intervisible, we need to find out presence of an obstacle-edge that blocks the straight line-of-sight joining $S_{in}$ and $D_{in}$. The brute-force approach is to check every obstacle-edge in $G_O$, for each $S_{in}$ and $D_{in}$ pair. But this induces a huge computational complexity, especially when the number of obstacles is huge e.g., in a metropolitan city or a disaster-affected landscape. Here, we propose a rotating sweep-line based approach where only a minimal subset of obstacle-edges need to be checked. Considering $S_{in}$ as the origin, we sort the vertices in $V$ = (set of unmarked obstacle-vertices in $G_O$ lying in the first- and fourth-quadrants for this shifted co-ordinate $\cup$ original intended destination $D$) in a decreasing order of slope. In situations where the slope is equal for two intermediate destinations, to break the tie we consider an increasing order of distance from source. Then, starting at the point with the the highest slope, we go on probing points in decreasing order of slope, with a clockwise-rotating sweep line pivoted at $S_{in}$. Further, we maintain a dynamic list $L_{cr}$ of \textit{critical-edges}. At each point probed, we add to $L_{cr}$ the set of clockwise edges attached to the point and exclude from $L_{cr}$ the set of anti-clockwise edges attached to the point. Thereby, the list of critical edges is never allowed to be large enough. For every new edge getting added in $L_{cr}$, we check whether it intersects the line joining $S_{in}$ and $D_{in}$. This is continued until the slope of the rotating sweep-line becomes less than the line joining $S_{in}$ and $D_{in}$. This is because, after that point there cannot be any edge that obstructs the line-of-sight between them. As such, $S_{in}$ and $D_{in}$ are intervisible if there are no critical edge that intersects their line-of-sight till this cut-off point.
\end{enumerate}

\noindent The Visibility graph $G_V$ is created by adding an edge between each pair of vertices whom we found to be intervisible by the previous steps. The corresponding edge-weight is the Euclidean distance of the line-of-sight between the vertices.

\paragraph{\textbf{Computing Shortest Path in Visibility Graph.}} The Visibility graph $G_V$ is effectively an Euclidean graph where intervisible vertices are joined by edges. For the pair of vertices that are mutually intervisible, the shortest path between them is the straight line-of-sight whose distance is the Euclidean distance. If two pair of vertices are not connected by an edge, we need a path of edges whose cumulative weight is minimum such that the traversed distance is shortest. Further, the source vertex $S$ is fixed because it is the position where the drone is currently located. The destination vertex $D$ is where the drone intends to go. So, we execute Dijkstra's Single-Source Shortest Path Algorithm on the Visibility graph $G_V$ to compute the shortest distance from $S$ to $D$. The drone can go along edges on the shortest path thus computed, because the edges indicate safe obstacle-free paths. Needless to say, the drone would have to change its heading angle of motion only when it transits from one edge to another. As shortest path indicates maximum stretches of straight line motions (as the ideal path i.e. the direct line-of-sight path is itself a single stretch of straight-line motion), it indicates lesser positions where the drone needs to maneuver mid-air steering. This effectively indicates that the time lost due to such maneuver is less and there are lesser chances of mid-air malfunctions due to sharp turns. 

\subsection{Long Journeys and Transformation to 3D Space}
\label{3dspace}

A drone is usually not intended for covering large distances as it has to make a round-trip journey to its base-station. Further, the battery consumption goes to the higher side when there are items that the drone has to carry. But, our approach can be extended to large distances by incorporating a trade-off between optimized and greedy. This can be done by keeping few \textit{stops} across the whole journey. When the drone reaches its first intermediate stop, it can repeat its maneuver of going to a high altitude and capturing RGB-D information. As such, it creates a new map layout with the first intermediate stop as its new source and continues its journey. In this case, the journey between intermediate stops is optimal and choice of stops is overall greedy. Further, in long journeys of the drone, even slow-moving obstacles can show substantial change in position between the start and end of flight of the drone. Such motions can be efficiently detected by unsupervised object tracking~\cite{jana2021unsupervised} on videos and the map layout can be dynamically modified on the run. 

It is known that finding shortest path in the 3D space and 3D path-planning is a NP-Hard problem~\cite{yang2014literature}. But 3D path-planning may be required when source and destination are at different altitudes. Our approach can be extended to 3D aerial space for drones through some heuristics. One approach is that the drone changes height (moves in z-axis) only at the stops and chooses a height that has least number of obstacle-grid-cells, through its perception subsystem. And at each height, it will use the proposed 2D optimized path-planning. Another alternative would be to consider multiple discretely rotated 2D planes (at small angles) pivoted at the source. The drone can then run the proposed path-planning technique on each plane and choose the plane which gives the least distance.

\section{Experimental Results}
\label{results}

The implementation included a simulation testbed for finding out the shortest obstacle-free path in a $m\times n$ rectangular grid with obstacles shown as shaded regions.

\begin{figure}
	\centering
	\begin{minipage}{0.32\textwidth}
		\centering
		\resizebox{\textwidth}{!}{%
			\begin{tikzpicture}
				\begin{axis}[
					xlabel={Grid-size ($g$) and \#Obstacles ($\frac{g^2}{5}$)},
					ylabel={Time in \textit{sec}},
					xmin=0, xmax=80,
					ymin=-0.1, ymax=25,
					xtick={5,10,20,30,40,50,60,70,80},
					ytick={0,5,10,15,20,25},
					xticklabel style={rotate=60},
					ymajorgrids=true,
					grid style=dashed,
					legend pos=north west]
					
					\addplot[
					style=ultra thick,
					color=red,
					mark=o,
					mark size=4pt
					] coordinates {(5,0.009)
						(10,0.027) 
						(20,0.076)
						(30,0.361)
						(40,1.135)
						(50,2.829)
						(60,6.907)
						(70,12.145)
						(80,22.809)};
					\addlegendentry{$g\times g$ grid, $\frac{g^2}{5}$ obstacles}
				\end{axis}
			\end{tikzpicture}
		}
		\label{fig:timeplot1}
	\end{minipage}
	\begin{minipage}{0.32\textwidth}
		\centering
		\resizebox{\textwidth}{!}{%
			\begin{tikzpicture}
				\begin{axis}[
					xlabel={Grid-size ($g$)},
					ylabel={Time in \textit{sec}},
					xmin=0, xmax=160,
					ymin=-1, ymax=1,
					xtick={5,10,20,40,80,160},
					ytick={-1,-0.5,0,0.5,1},
					xticklabel style={rotate=60},
					ymajorgrids=true,
					grid style=dashed,
					legend pos=north west]
					
					\addplot[
					style=ultra thick,
					color=blue,
					mark=square,
					mark size=4pt
					] coordinates {(5,0.061)
						(10,0.024) 
						(20,0.039)
						(40,0.032)
						(80,0.037)
						(160,0.043)};
					\addlegendentry{$g\times g$ grid, $20$ obstacles}
				\end{axis}
			\end{tikzpicture}
		}
		\label{fig:timeplot2}
	\end{minipage}
	\begin{minipage}{0.32\textwidth}
		\centering
		\resizebox{\textwidth}{!}{%
			\begin{tikzpicture}
				\begin{axis}[
					xlabel={\#Obstacles ($o$)},
					ylabel={Time in \textit{sec}},
					xmin=40, xmax=1280,
					ymin=0, ymax=25,
					xtick={40,80,160,320,640,1280},
					ytick={0,5,10,15,20,25},
					xticklabel style={rotate=60},
					ymajorgrids=true,
					grid style=dashed,
					legend pos=north west]
					
					\addplot[
					style=ultra thick,
					color=magenta,
					mark=diamond,
					mark size=5pt
					] coordinates {(40,0.063)
						(80,0.208) 
						(160,0.632)
						(320,2.002)
						(640,6.663)
						(1280,22.809)};
					\addlegendentry{$80\times 80$ grid, $o$ obstacles}
				\end{axis}
			\end{tikzpicture}
		}
		\label{fig:timeplot3}
	\end{minipage}
	\caption{Variation in execution-time with (a) simultaneous change in grid-size and number of obstacles, where obstacle occupancy is 20\% (b) change in grid-size when number of obstacles is fixed (c) change in number of obstacles, with fixed grid-size}
	\label{fig:timePlots}
\end{figure}
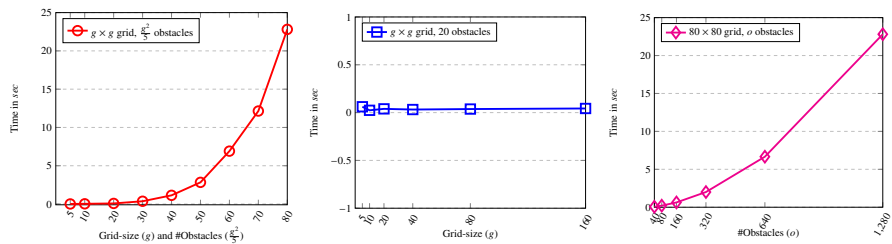

\noindent Obstacles are square-shaped blocks denoted by shaded grid-cells. The shortest path is shown as a red line. It may be noted that drones are of insignificant size w.r.t. obstacles, so drones may travel through boundaries between two separate objects that contain sufficient space in between. The source code is written in C++~\cite{jana2014c++} on Ubuntu platform.

While running the simulation experiments for shortest obstacle-free path, we measure the elapsed execution time by varying grid-size, number of obstacles and observe the effect in Fig.~\ref{fig:timePlots}. With simultaneous change of grid-size and number of obstacles in Fig.~\ref{fig:timePlots}(a), we observe that as expected, when grid size and number of obstacles increase, the execution time increases exponentially. Keeping the number of obstacles fixed in Fig.~\ref{fig:timePlots}(b), as we increase grid-size, we see time does not vary much. Thus, we may infer that execution time depends on number of obstacles and not much on grid-size. This can be well illustrated in Fig.~\ref{fig:timePlots}(c) where execution time varies almost linearly with varying number of obstacles keeping the grid-size same.

\begin{figure}[!htb]
	\centering
	\begin{minipage}{.37\textwidth}
		\centering
		\begin{subfigure}{0.9\linewidth}
		\centering
		\fbox{
		\includegraphics[width=\linewidth]{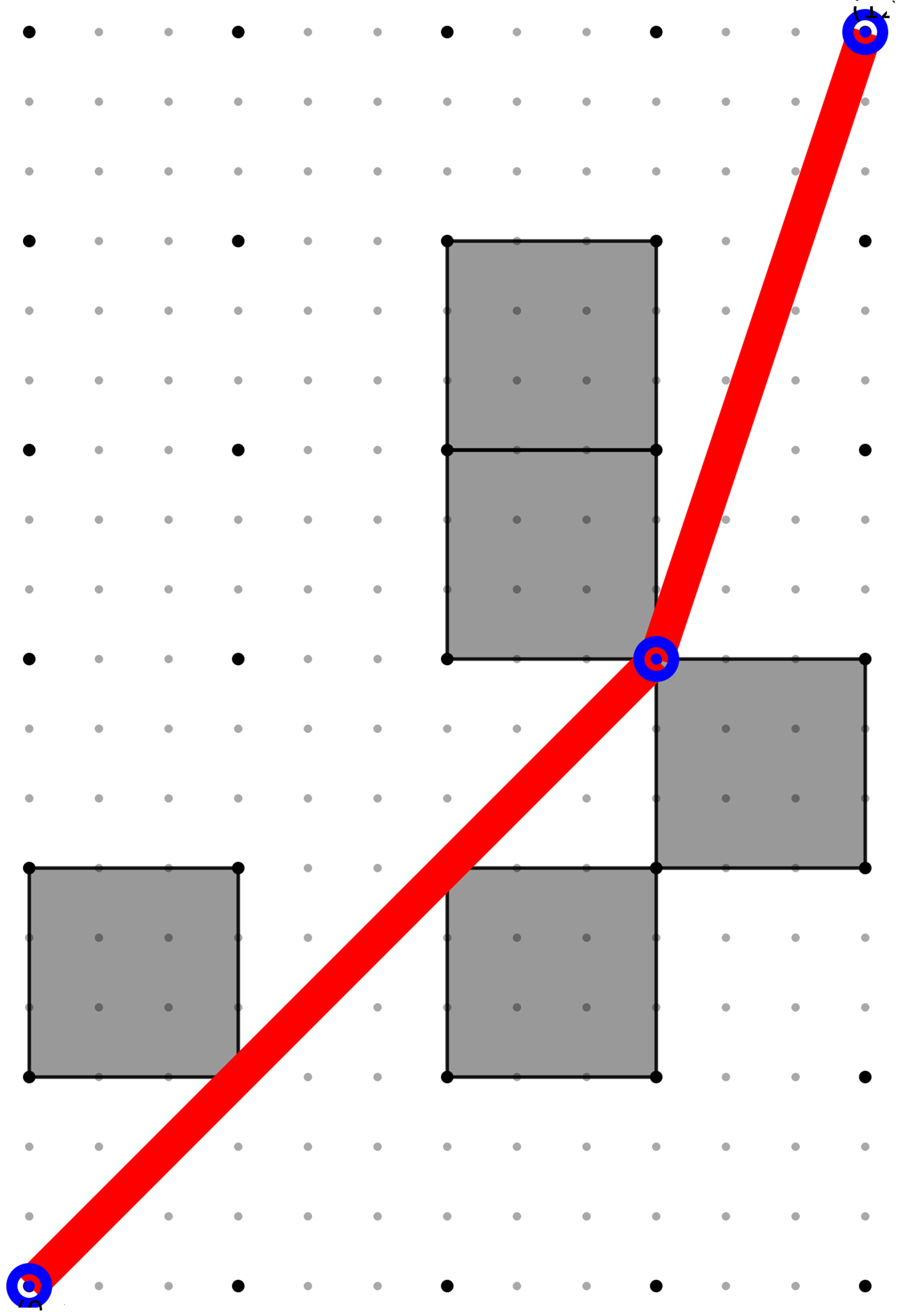}  }
		\caption{}
		\label{fig:spathResult1_elementary}
		\end{subfigure}
		\\
		\vspace{0.045\textwidth}%
		\begin{subfigure}{0.93\linewidth}
		\centering
		\fbox{
		\includegraphics[width=\linewidth]{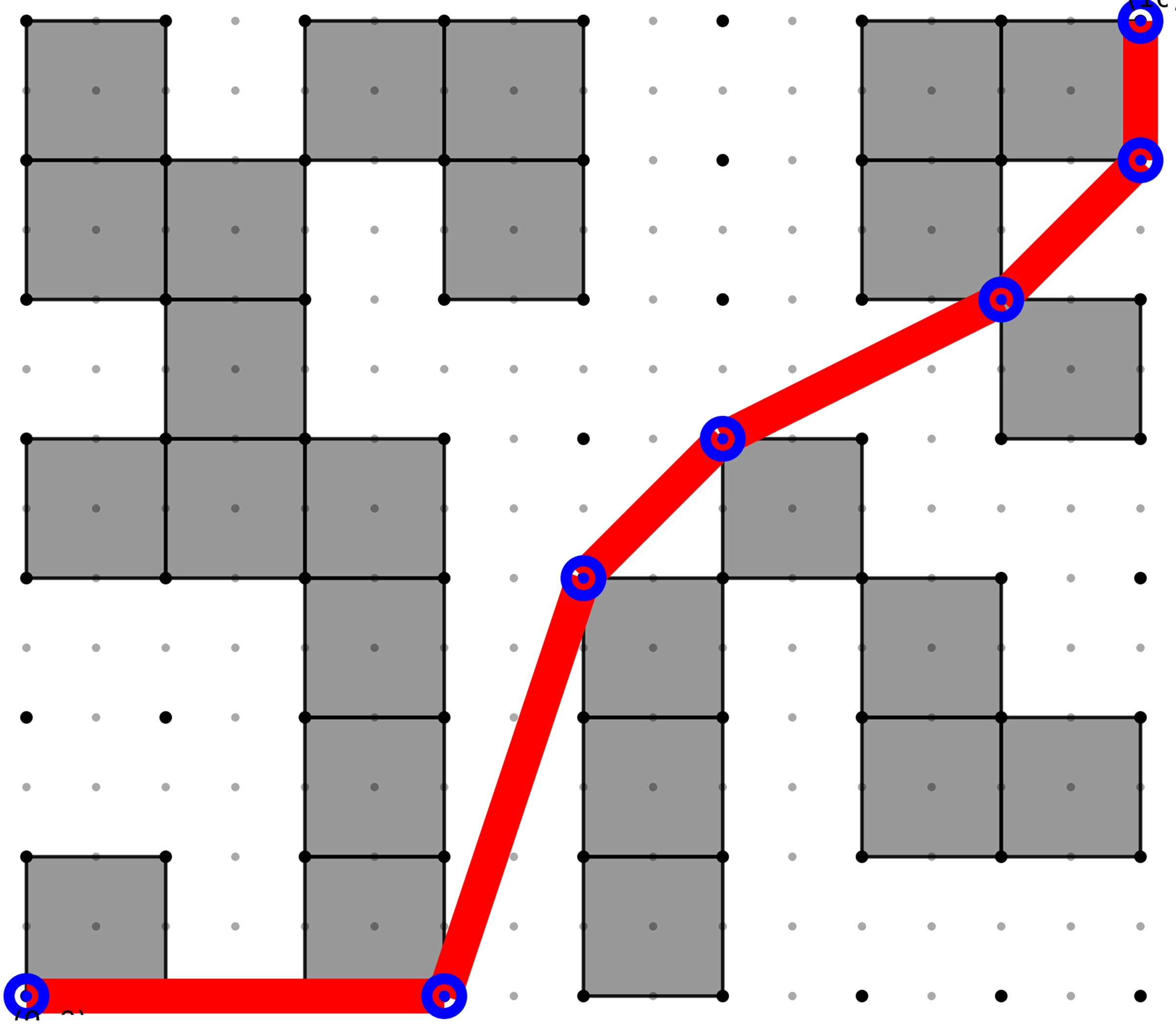}}  
		\caption{}
		\label{fig:spathResult2_small}
		\end{subfigure}
	\end{minipage}%
	\hspace{0.02\textwidth}%
	\begin{minipage}{0.59\textwidth}
		\centering
		\begin{subfigure}{\linewidth}
		\centering
		\fbox{
		\includegraphics[width=\linewidth]{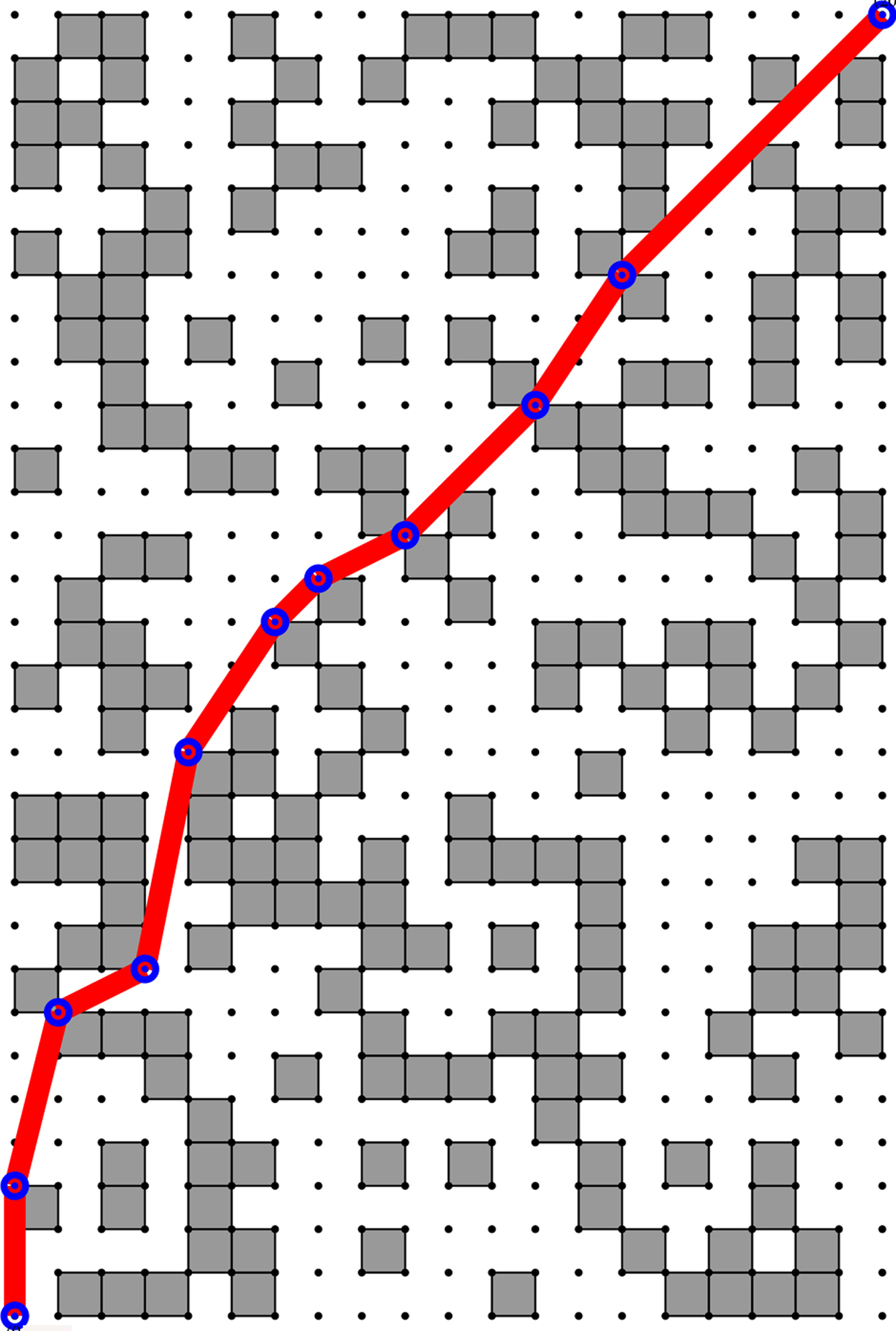}}
		\caption{}
		\label{fig:spathResult3_medium}
		\end{subfigure}
	\end{minipage}
\caption{Simulations on (a-b) small- and (c) medium-scale maps. Obstacles are shown as shaded squares, shortest safe path as red line and deflection points as blue circles}
\label{fig:spathResult_small_medium}
\end{figure}

\newpage

\begin{figure}[!h]
	\centering
	\fbox{
	\includegraphics[width=0.84\textwidth]{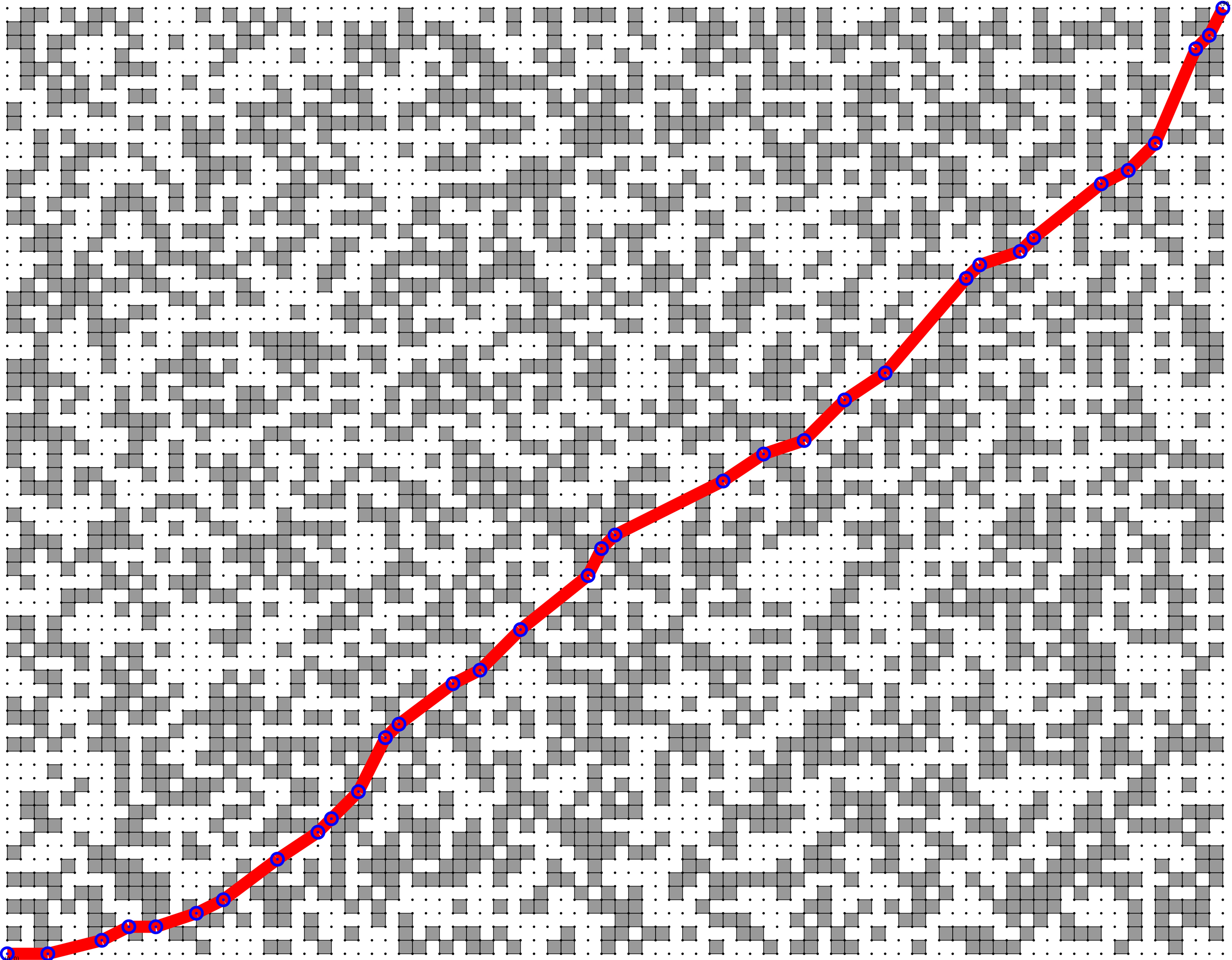}}
	\caption{A $70\times 90$ rectangular grid with each cell of size $1\times 1$, and $2500$ obstacles}
	\label{fig:spathResult4_large}
\end{figure}


\begin{figure}[!h]
	\centering
	\fbox{
	\includegraphics[width=0.84\textwidth]{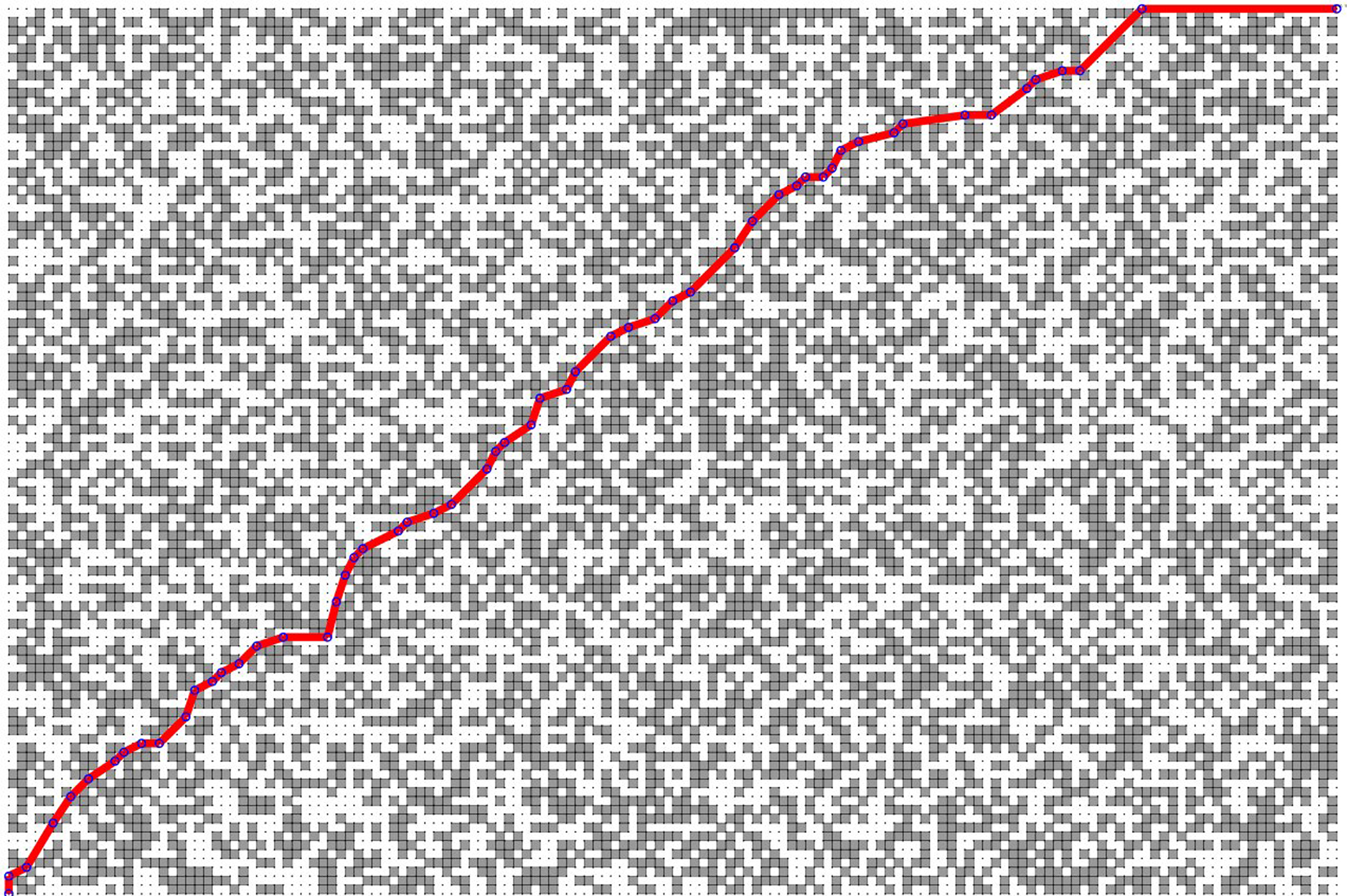}}
	\caption{A $100\times 150$ rectangular grid with each cell of size $1\times 1$, and $7000$ obstacles}
	\label{fig:spathResult5_larger}
\end{figure}


In Figures~\ref{fig:spathResult_small_medium}-\ref{fig:spathResult5_larger}, we present some illustrations of the simulations done on different grid-sizes for proportionally increasing number of obstacles. In Figure~\ref{fig:spathResult1_elementary}, there are $5$ obstacles spread over a $6\times 4$ rectangular grid, each cell being of size $3 \text{ units}\times 3\text{ units}$. The shortest path is of length $\sqrt{162}+\sqrt{90}=22.21$ units. On a similar note, Figure~\ref{fig:spathResult2_small} shows a $7\times 8$ rectangular grid with each cell of size $2 \text{ units}\times 2\text{ units}$, and $25$ obstacles. Shortest path is of length $\sqrt{36}+\sqrt{40}+\sqrt{8}+\sqrt{20}+\sqrt{8}+\sqrt{4}=24.45$ units. Figure~\ref{fig:spathResult3_medium} is a medium-scale grid of $30\times 20$ with $200$ obstacles and each cell is of unit size. Here, the shortest path is of length $\sqrt{9}+\sqrt{17}+\sqrt{5}+\sqrt{26}+\sqrt{13}+\sqrt{2}+\sqrt{5}+\sqrt{18}+\sqrt{13}+\sqrt{72}=38.05$ units.

Figures~\ref{fig:spathResult4_large}-\ref{fig:spathResult5_larger} are examples of simulation on even larger grid-sizes with more number of obstacles. While Figure~\ref{fig:spathResult4_large} shows application of the proposed method on a grid of size $70\times 90$ with $2500$ obstacles ($\sim$40\% obstacle occupancy), Figure~\ref{fig:spathResult5_larger} shows it on a $100\times 150$ grid with $7000$ obstacles ($\sim$47\% obstacle occupancy). The shortest path lengths are $ 118.53 $ units and $ 195.39 $ units respectively.

\section{Conclusion and Future Scope}
\label{conclusion}
In this work, we have achieved finding the shortest obstacle-free path from a given source-node to a designated target-node via multiple node hops, through an unobstructed flight-route with clear visibility. AutoDrones following this technique will encounter less number of angular direction changes while attempting to fly in this route. This approach has many societal benefits of utilizing AutoDrones for reaching target in shortest aviation path, thus becoming energy-efficient too. Through perception subsystem, the proposed approach refrains from using any preset maps and instead, constructs dynamic maps during travel. We have presumed that all the obstacles are fully-static like buildings, monuments etc. or at least static till the drone again computes dynamic route at its intermediate stop. However, in reality, there could be moving obstacles in the flyway when the drone is flying at a low altitude, like walking humans, another drone or flying bird. The computation of the next move could be more dynamic in nature to tackle such cases, which we aim to address in our future works.

\bibliographystyle{spmpsci}
\bibliography{references}

\end{document}